\documentclass[twocolumn, final]{elsarticle}

\usepackage{framed,multirow}

\usepackage{amssymb}
\usepackage{latexsym}

\usepackage{xcolor}
\definecolor{newcolor}{rgb}{.8,.349,.1}

\usepackage{amsfonts}
\usepackage{graphicx}
\usepackage{bbding}

\usepackage{braket}
\usepackage{amsmath}
\usepackage{color}
\usepackage[utf8]{inputenc}
\usepackage{cases}
\usepackage{nicefrac}
\usepackage{tabu}
\usepackage{booktabs}
\usepackage{multirow}
\usepackage{subcaption}
\usepackage{caption}
\usepackage{soul}
\usepackage{kotex}
\usepackage{arydshln}
\usepackage[linesnumbered,ruled]{algorithm2e}

\setstcolor{red}

\begin{document}

\clearpage

\ifpreprint
  \setcounter{page}{1}
\else
  \setcounter{page}{1}
\fi

\begin{frontmatter}

\title{Arbitrary-Scale Downscaling of Tidal Current Data Using Implicit Continuous Representation}

\cortext[cor]{Corresponding author: youngmin.ro@uos.ac.kr (Youngmin Ro)\\
e-mails: dslisleedh@uos.ac.kr, jeongsm@ust21.co.kr
}

% \author[1]{Dongheon \snm{Lee}}   #### Commented. 0130 for no style submission
% \author[2]{Seungmyong \snm{Jeong}}
% \author[1]{Youngmin \snm{Ro}\corref{cor}}
\author[1]{Dongheon Lee}
\author[2]{Seungmyong Jeong}
\author[1]{Youngmin Ro\corref{cor}}

\address[1]{Department of Artificial Intelligence, University of Seoul, 163, Seoulsiripdae-ro, Dongdaemun-gu, Seoul, Korea 02504}
\address[2]{UST21, 129, Gaetbeol-ro, Yeonsu-gu, Incheon, Korea 21999}
% \address[3]{Department of Advanced Imaging, Chung-Ang University, 84 Heukseok-ro, Dongjak-gu, Seoul, Korea 06974}
% \address[5]{Samsung SDS, 125, Olympic-ro 35-gil, Songpa-gu, Seoul, Korea 05510}

\begin{abstract}
Numerical models have long been used to understand geoscientific phenomena, including tidal currents, crucial for renewable energy production and coastal engineering. However, their computational cost hinders generating data of varying resolutions. 
As an alternative, deep learning-based downscaling methods have gained traction due to their faster inference speeds.
But most of them are limited to only inference fixed scale and overlook important characteristics of target geoscientific data.
In this paper, we propose a novel downscaling framework for tidal current data, addressing its unique characteristics, which are dissimilar to images: heterogeneity and local dependency. 
Moreover, our framework can generate any arbitrary-scale output utilizing a continuous representation model.
Our proposed framework demonstrates significantly improved flow velocity predictions by 93.21\% (MSE) and 63.85\% (MAE) compared to the Baseline model while achieving a remarkable 33.2\% reduction in FLOPs. 
\end{abstract}

% \begin{keyword}
% \KWD Arbitrary-scale downscaling\sep Image Super-Resolution\sep Implicit Neural Representation\sep Oceanic Tidal Current Data
% Arbitrary-scale downscaling\sep Image Super-Resolution\sep Implicit Neural Representation\sep Oceanic Tidal Current Data
% \end{keyword}

\end{frontmatter}

%\linenumbers

% \input{1-Introduction}

\section{Introduction}
Even before deep learning became mainstream, there were many efforts to understand and predict weather phenomena using numerical modeling, which mathematically represents complex physical phenomena.
Numerical model simulation data on tidal currents, an essential component in understanding and predicting natural phenomena, is highly effective in capturing flow patterns and aspects within a specific region based on partially observed real-time data. 
Furthermore, the accurate future prediction of tidal currents is a crucial stage for renewable energy production~\cite{nachtane2020tidalcurrent_turbine}. 
The information on tidal currents is also utilized in coastal engineering, which is based on the interaction of tides with offshore structures~\cite{thieler2000coastalengineering}. 
Different resolutions of tidal current data are needed for accurate decision-making and applying various types of research. 
However, the computational cost of the numerical models is significant: it requires a vast number of CPU cores and grows with the number of pixels being calculated.
Therefore, it is prohibitively expensive to use numerical models to generate tidal current data of varying resolution over large areas.
As an alternative, downscaling methods~\cite{giorgi1999dynamicaldownscaling}, which use hierarchical models to generate data of different resolutions, are employed. Initially, a main model is used to generate low-resolution data over a broad area. Subsequently, various sub-models are used to generate arbitrary-resolution data for nest areas of interest.

\begin{figure}
    \centering
    \includegraphics[width=1\linewidth]{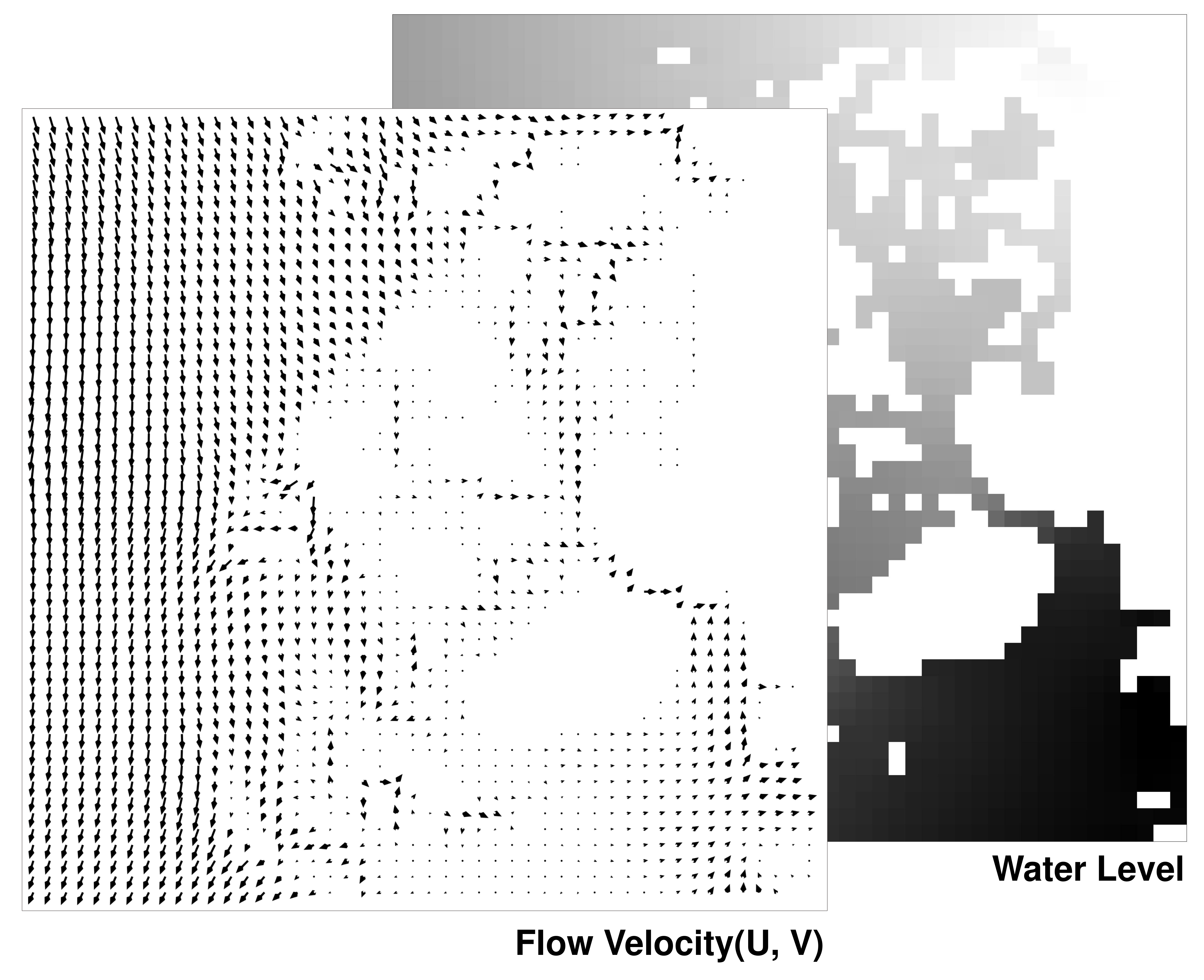}
    \caption{\textbf{Visualization of Tidal Current Data.} 
    Unlike images, tidal current data consists of two channels, U and V, representing the flow velocity, and one channel representing the water level.
    Both the features are highly correlated but show very different aspects.
    }\label{fig:inputs_gray}
\end{figure}

Recently, there has been growing interest in exploring the use of deep learning models for downscaling~\cite{geiss2022downscaling, harris2022generative} due to their considerably faster inference speeds than numerical models. 
Many researchers generally used deep learning models from the image super-resolution domain where the image's resolution is enhanced for better image quality.
However, as shown in Figure~\ref{fig:inputs_gray}, tidal current data possesses unique characteristics that differentiate it from images in the following three aspects. 
1) Firstly, although tidal current data shares a three-channel tensor structure similar to images, it comprises information with different characteristics than images.
This implies that the pre-trained parameters on large image datasets such as ImageNet~\cite{deng2009imagenet} or {DIV2K~\cite{Agustsson_2017_CVPRW_DIV2K} could not be utilized as the general super-resolution task does~\cite{chen2021pre_CVPR, chen2023hat_CVPR, zhang2022swinfir}.
2) Secondly, in the tidal current data, the first two channels represent the flow velocity (U and V vectors), and the third channel represents the water level. 
Water level refers to the vertical difference caused by the gravitational interaction between the moon, sun, and earth, and flow velocity refers to the magnitude of horizontal movement resulting from the gradient of water levels.
Even though these are highly correlated, their characteristics are distinct and inherently heterogeneous.
3) Lastly, each grid in tidal current data maps to a specific geographical region. 
This means, unlike images super-resolution domain that assume translation invariance, consideration needs to be given to where certain patterns appear.

\begin{figure*}[t!]
  \centering
  \includegraphics[width=0.8\linewidth]{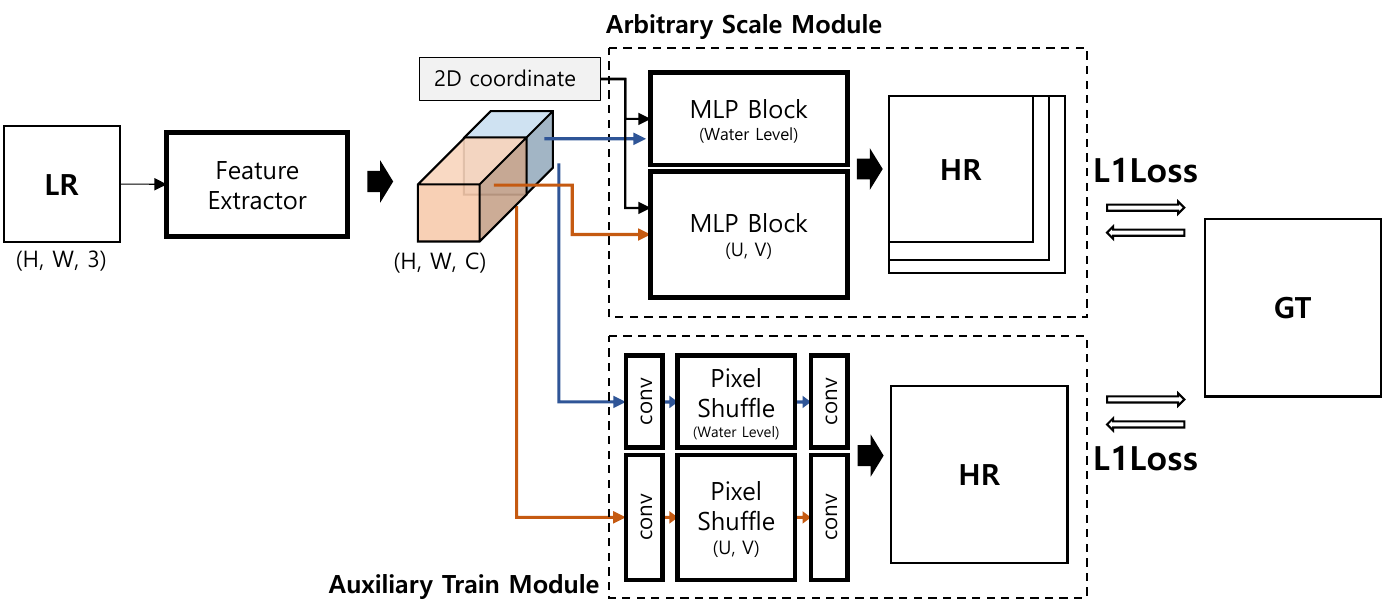}
  \vspace{-0.3cm}
  \caption{
  The overall framework of the proposed downscaling for tidal current data. 
  Our framework consists of three parts: a feature extractor to extract features from low-resolution input, an arbitrary-scale module to predict arbitrary-scale downscaling prediction, and an auxiliary train module added for faster convergence and to prevent over-fitting of the feature extractor. Each output from the arbitrary-scale module and the auxiliary train module generates gradients and is used to train the feature extractor
  }
  \label{fig:overall_scheme}
  \vspace{-0.3cm}
\end{figure*}

In this paper, we propose a novel downscaling framework that is effectively applied to tidal current data.  
To this end, we design our new framework with the following three considerations.
1) Since the unavailability of a pre-trained feature extractor, we propose an Auxiliary Train Module (ATM) that can rapidly and stably train a feature extractor from scratch. 
2) Considering the heterogeneous nature of flow velocity and water level, we suggest a Feature Map Splitting (FMS) to disentangle and up-sample them separately.
In addition, the FMS effectively reduces the extra computation cost caused by adding the ATM during the training process.
3) We insert a Positional Encoding (PE) layer into the feature extractor to take advantage of geographical region cues.

In our validation experiments, the proposed framework is shown to be successfully applied to tidal current data based on a model~\cite{LIIF_Chen_2021_CVPR} (Baseline), which utilized continuous representation that was originally used in the 3D reconstruction domain into image super-resolution domain.
Compared to the Baseline, our framework shows not only improving performance on flow velocity by 93.21\% and 63.85\% for MSE and MAE, respectively but also reduces FLOPs by 33.2\%.
In addition, the ablation study of each module shows a comprehensive evaluation of each component of our proposal.
Through comparison of visualization, we show that our model downscales the out-of-train distribution more effectively than traditional methods such as Bicubic and the Baseline. 

\section{Related Work}
Numerous studies~\cite{vandal2017deepsd, groenke2020climalign, LesLap_2020, MSG-GAN-SD_2021, ClimaX_2023} have utilized deep learning for downscaling numerical model simulation data.
ResLap~\cite{LesLap_2020} employed LapSRN~\cite{LapSRN_2017}, a model that proposes progressive upsampling as its backbone and calculated losses from each progressively downscaled predictions using a root mean square error.
MSG-GAN-SD~\cite{MSG-GAN-SD_2021} used all progressively downscaled predictions for a GAN (Generative Adversarial Networks) discriminator, to leverage semantic information appearing at different scales.
While both studies' models could generate outputs at various scales, they could not infer at scales not used during training. Addressing this, we leverage an arbitrary-scale image super-resolution model~\cite{LIIF_Chen_2021_CVPR} as a Baseline. 
ClimaX~\cite{ClimaX_2023} aimed to create a geospatial-temporal foundation model using the Transformer~\cite{ViT_2021} as a feature extractor and then downstream it to various tasks, including downscaling. 
However, numerical model simulation data varies widely in spatial size, so it is computationally expensive to use the transformer, which has a quadratic computational cost as the spatial size of the input increases.

In the 3D Reconstruction domain, there have been many successful works employing INR-based models, aiming to learn an implicit continuous representation of partially observed images~\cite{mildenhall2021nerf, mescheder2019occupancy, sitzmann2020siren}.
LIIF~\cite{LIIF_Chen_2021_CVPR} adopted the INR-based model for the arbitrary-scale image super-resolution domain, outperforming pre-existing arbitrary-scale image super-resolution models~\cite{hu2019metasr}.
Since that time, various researchers have attempted to use INR-based models for arbitrary-scale image super-resolution~\cite{lee2022lte, pak2023btc, cao2023ciaosr}.
But in the downscaling task, which utilizes models and methods from image super-resolution, no research uses the INR-based model as far as we know. 
Therefore, in this paper, we introduce the INR-based model in the downscaling task and modify it to suit the characteristics of tidal current data, rather than simply using it straightforwardly.

\section{Methodology}
\subsection{Overview}
In Figure.~\ref{fig:overall_scheme}, we show the overall framework of the proposed arbitrary-scale downscaling model for tidal current data. 
The proposed model consists of three parts. 
First, the feature extractor takes low-resolution input data and outputs a feature map.
Second, the Arbitrary Scale Module (ASM) takes the feature map and relative 2D coordinates and then predicts the arbitrary-scale output (HR$^{arb}$) using MLP blocks. 
Last, the Auxiliary Train Module (ATM) takes the feature map and predicts fixed-scale high-resolution output (HR$^{fixed}$) using a convolution layer and pixel shuffle transformation. 
The ASM and the ATM predict flow velocity and water level separately by splitting the feature map into two parts according to the Feature Map Splitting (FMS) ratio. 
The details of the modules will be explained in the following sections.

\subsection{Arbitrary Scale Module}
To predict arbitrary-scale downscaled HR (HR$^{arb}$), We utilize the INR-based module consisting of simple MLP (Multi-Layered Perceptrons) blocks. The way to generate HR$^{arb}$ using ATM is formulated as Eq.~\ref{eq::liif}.
\begin{equation}\label{eq::liif}
    \hat{s} = f_\theta(z, x).
\end{equation} 
The MLP blocks $f$ with parameters $\theta$ predicts the signal $\hat{s}$ at a specific point using the feature vector $z$ extracted from the LR by the feature extractor and a 2D coordinate $x$ representing the relative location to be predicted.
In the context of image super-resolution tasks, the signal corresponds to the RGB values, while in this study, it corresponds to the flow velocity and water level.
By training the MLP blocks to predict output given feature map and relative 2D coordinate, it learns an implicit continuous representation of the input. This allows the ASM module to produce high-resolution output at any desired scale just sampling the related 2D coordinates densely.

\subsection{Auxiliary Train Module}
In the image super-resolution domain, the feature extractor is pre-trained on an image dataset such as DIV2K~\cite{Agustsson_2017_CVPRW_DIV2K}.
However, tidal current data has different characteristic from the natural images, so the pre-trained parameters cannot be used as it is.
This yields the optimization difficulty of training both the feature extractor and ASM from scratch.
To solve the above problems, we present the auxiliary train module (ATM), which helps to reliably and quickly train feature extractors without any pre-trained parameters.
In the ATM, the simple 3$\times$3 convolution layers and pixel shuffle transformation~\cite{espcn_2016} are used. 
The first convolutional operation is employed to amplify the channel size, followed by a pixel shuffle operation, which transfers channel dimensions into spatial dimensions. 
The final convolutional operation is then used to align the channel dimensions with the target's channel dimensions. 
Since the ATM doesn't use coordinate sampling like the ASM, gradients calculated from HR$^{fixed}$ are more structured than gradients from HR$^{arb}$.

\subsection{Feature Map Split downscaling} \label{subsec::feature_split_upsampling}
In our framework, to separately predict the distinct properties of flow velocity and water level, we propose Feature Map Split (FMS) downscaling, which splits the feature map produced by the feature extractor at a pre-defined ratio.
Then, each part downscales flow velocity and water level separately from the given feature map through different convolutional and MLP layers.
Channels and units in downscaling modules are also split according to the FMS ratio.
We consider several FMS ratios such as \{None, 2:1, 5:1, 11:1\}. 
The proposed FMS not only has the ability to separate outputs with different characteristics but also contributes to building an efficient framework by reducing computational costs.
As shown in Table~\ref{table:flops_comparison}, FMS reduces the computational costs by 34\% and 33\% in train and test, respectively, 
Our model with a 2:1 ratio requires 611G FLOPs at train time despite applying an additional module (ATM). 
This is less than the 819G FLOPs of computation over the test time of a model without FMS. 
Additionally, the performance is also improved when FMS is applied, as shown in Table~\ref{table::split_eval}. 

\renewcommand{\arraystretch}{} 
\begin{table}[t]
\centering
\caption{
Computation costs according to the proposed Feature Map Splitting (FMS) ratio. 
As the FMS ratio becomes more balanced, the computational cost decreases. At a 2:1 ratio, it can be seen that even the train state (w/ the ATM) has less computational cost than computational cost w/o both the ATM and the FMS.
}
\resizebox{0.8\columnwidth}{!}{
\begin{tabular}{@{}cccccc@{}}
\toprule
\multirow{2}{*}{FMS} & \multicolumn{5}{c}{FLOPs (G)}            \\ \cmidrule(l){2-6} 
                     & FE  & ASM & \textbf{Test} & +ATM & \textbf{Train} \\ \midrule
None                 & 207 & 612 & 819           & +116 & 934            \\
11:1                 & 207 & 514 & 721           & +96  & 817            \\
5:1                  & 207 & 442 & 649           & +83  & 732            \\
2:1(Ours)            & 207 & 340 & \textbf{547}  & +64  & \textbf{611}   \\ \bottomrule
\end{tabular}
}

\label{table:flops_comparison}
\end{table}

\subsection{Feature Extractor} \label{subsec::feature_extractor}
We use a CNN-based feature extractor~\cite{EDSR_2017} and insert a Positional Encoding (PE) layer into the feature extractor to utilize the translation variant geographical cues of tidal current data.
Among the several types of PE~\cite{Transformer_2017, ViT_2021, baevski2020wav2vec_cpe, chu2021conditional}, we utilized learnable PE, which increases negligible computational load and is easy to implement.
It is implemented by adding a learnable parameter of the same size to the feature map that passes through the first convolution layer. 
Since we do not randomly crop or flip an input when training the model, each grid is always added to the same positional encoding.
The structure of the feature extractor is illustrated in Fig.~\ref{fig:feat_ext_arc}.

\begin{figure}
\centering
    \includegraphics[width=0.8\linewidth]{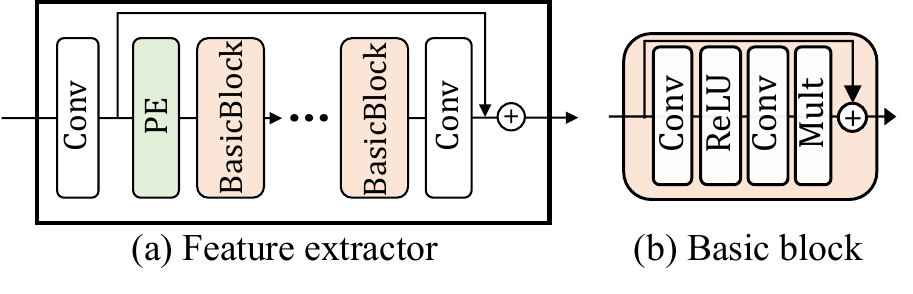}
    \caption{
        Feature extractor architecture.
        We use EDSR~\cite{EDSR_2017} as a feature extractor and add learnable positional encoding before the first basic block.
    }\label{fig:feat_ext_arc}
\end{figure}

\section{Experiments}
\subsection{Datasets}
The data used for training are tidal current data produced by KIOST(Korea Institute of Ocean Science \& Technology) using the MOHID water modeling system. 
The data consists of pairs, each including information generated on a 2km spatial grid with $50\times 48$ size and a 300m spatial grid with $300\times 288$ size within the range of latitude 34.1458 E to 35.1875 E, longitude 125.5416 N to 126.5416 N from 2020/01/01 to 2023/01/01. 

\subsection{Preprocessing}
The land portion of the data range was filled with an invalid value, so we masked it and interpolated the value with the nearest valid pixel value because we were concerned that interpolating with a specific constant (e.g., zero) would affect the nearby invalid pixels in the convolution operation. 
In the Baseline~\cite{LIIF_Chen_2021_CVPR}, they first crop patch from HR according to random scale (from $\times$1 to $\times$4) multiplied by pre-defined low-resolution input size and then use it as a GT (target). LR input was generated by applying bicubic downsampling to the cropped GT according to the random scale used before.
However, this approach is not entirely appropriate for tidal current data for a couple of reasons. 
1) Each grid in the tidal current data represents a specific region. 
Training with randomly cropped LR/GT pairs can make it more challenging to learn the corresponding regional information.
2) The tidal current used for training consists of pre-existing LR and GT pairs, each generated by its own numerical model. 
Resizing the LR or GT using interpolation methods such as bicubic to create a random scale LR input, as done in the original method, is not the most suitable approach for tidal current data. Therefore, in our approach, we extracted the $x$ and $s$ from GT without any transformation to them, and we used the LR as it is for the input. 
It was reported in the previous research~\cite{LIIF_Chen_2021_CVPR} that models trained only at $\times$4 scale showed higher PSNR at the out-of-train distribution scale(greater than $\times$4) compared to models trained on random scales. And since we using a fixed-scale LR-HR pair, we did not use cell decoding proposed in previous research.

\subsection{Implementation details}
The model is implemented using Pytorch and trained using a single RTX3090. 
The learnable positional encoding is initialized from Gaussian distribution $\mathcal{N}(0,~0.02)$. 
The number of filters in the model is set to 384, and the number of basic blocks is set to 32. 
We use 4 MLP blocks inside of the ASM and each MLP block consists of the MLP layer with 384 units and ReLU activation except for the last layer which consists of a single MLP layer. 
Units in MLP can be split according to the FMS ratio.

We trained the model for 50 epochs with Adam Optimizer. 
The learning rate is initially set to 0.0001 and reduced by a factor of 10 at the 25th epoch. 
Both modules trained using L1 loss, and areas corresponding to land are masked and excluded from the loss calculation.

\subsection{Results}

\subsubsection{Ablation Study}\label{subsubsec:ablation_study}
To evaluate the model, mean squared error (MSE, Eq.~\ref{eq::MSE}) and mean absolute error (MAE, Eq.~\ref{eq::MAE}) between the test ground truth data and HR$^{arb}$ with an in-train distribution of $\times$6 scale are measured. 
\begin{equation}\label{eq::MSE}
    MSE = \frac{1}{n}\sum_{i=1}^{n}(gt_i-hr^{arb}_i)^2
\end{equation} 
\begin{equation}\label{eq::MAE}
    MAE = \frac{1}{n}\sum_{i=1}^{n}|gt_i-hr^{arb}_i|
\end{equation} 
Flow velocity (u, v) and water level are evaluated separately, in line with the water modeling system~\cite{blain1998coastal, gomes2015tidemohid}.
We conducted an ablation study to assess the effectiveness of the proposed module and components such as the ATM, the PE, and the FMS. 
The first row of Table~\ref{table::auxpe_eval} shows the Baseline, which means the result of applying the existing super-resolution model straightforwardly to the tidal current data. 
As shown in the second row, adding the ATM to the Baseline significantly enhances the prediction performance for flow velocity, which is more challenging to predict. 
This result confirms that ATM effectively contributes to learning the feature extractor. 
Further, Fig.~\ref{fig:atm_ablation} shows train losses and validation MSE whether w/ and w/o the ATM. 
In the early train phase, loss w/ the ATM is lower than loss w/o the ATM which means the ATM helps to train the feature extractor faster. 
However, in the late train phase, loss w/o the ATM is lower than w/ the ATM, but validation MSE was not, which means the ATM prevents the feature extractor from over-fitting.
As shown in the third row of Table~\ref{table::auxpe_eval}, when we add the PE, it is shown that the performance continues to increase. 
This means that utilizing cues from geographical regions with the PE is appropriate.
The last row shows that applying FMS improves performance on both MSE and MAE.
This confirms that splitting the feature map and predicting each feature is very helpful. 

\begin{table}[t]
\centering
\caption{
Ablation on the Proposed Methods. 
Each metric is rescaled according to the units indicated. 
The performance improves with each addition of the proposed method.
}
% \captionsetup{justification=centering}
\resizebox{0.9\columnwidth}{!}{
\begin{tabular}{@{}ccccccc@{}}
\toprule
\multirow{3}{*}{ATM} & \multirow{3}{*}{PE} & \multirow{3}{*}{\begin{tabular}[c]{@{}c@{}}FMS\\ (2:1)\end{tabular}} & \multicolumn{4}{c}{Arb. Scale Module}                           \\ \cmidrule(l){4-7} 
                     &                     &                                                                      & \multicolumn{2}{c}{MSE ($\times$1e-5)} & \multicolumn{2}{c}{MAE ($\times$1e-3)} \\ \cmidrule(l){4-7} 
                     &                     &                                                                      & Velocity        & Level        & Velocity        & Level        \\ \midrule
\XSolidBrush                    & \XSolidBrush                   & \XSolidBrush                                                                    & 18.488          & 0.276        & 3.879           & 1.076        \\
\CheckmarkBold                    & \XSolidBrush                   & \XSolidBrush                                                                    & 4.980           & 0.250        & 2.108           & 1.194        \\
\CheckmarkBold                    & \CheckmarkBold                   & \XSolidBrush                                                                    & 3.322           & 0.202        & 3.032           & 1.049        \\
\CheckmarkBold                    & \CheckmarkBold                   & \CheckmarkBold                                                                    & \textbf{1.255}           & \textbf{0.144}        & \textbf{1.403}           & \textbf{0.961}        \\ \bottomrule
\end{tabular}
}
\label{table::auxpe_eval}
\end{table}
\begin{figure}
\centering
    \includegraphics[width=1.\linewidth]{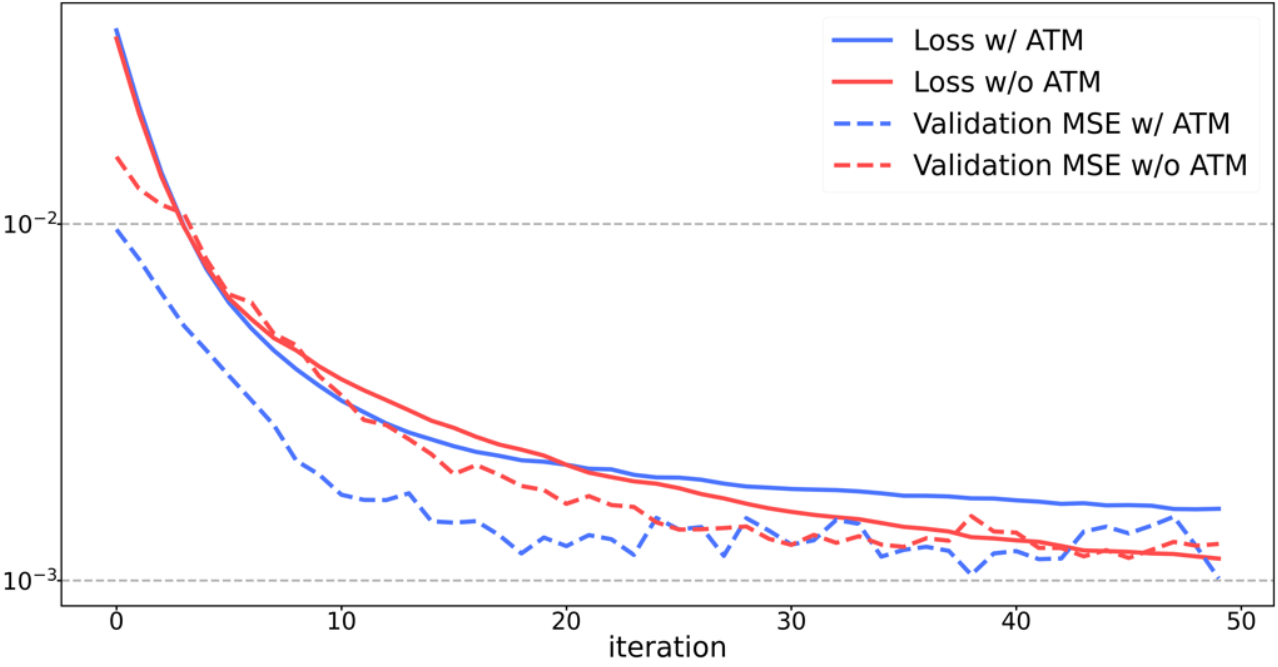}
    \caption{
        Train losses and validation MSE comparison between w/ and w/o the ATM. Both train losses and validation MSE are calculated between HR$^{arb}$ and GT.
    }\label{fig:atm_ablation}
\end{figure}

\subsubsection{Trade-offs with Varying FMS Ratios}
In this section, the experiments are conducted to examine the trade-off between computational cost and performance depending on the FMS ratio.
We consider four variants of FMS: \{None, 2:1, 5:1, 11:1\} and measure the computational cost and performance.
In Table~\ref{table::split_eval}, the MSE performance of the 2:1 FMS variant is higher on average than the MSE performance of the 11:1 FMS variant. 
On the contrary, the computational cost of the 2:1 FMS variant is lower than that of the 11:1 FMS variant.
In other words, the 2:1 variant shows a benefit in terms of computational cost, the 11:1 variant shows a benefit in performance, and thus the split ratio can be set according to the user's needs. 
Interestingly, the 5:1 variant shows lower performance than the 2:1 and 11:1 variants. 
This suggests that, unlike computational cost, there is not a proportional trade-off in performance based on the FMS ratio.

\begin{table}[t]
\centering
\caption{Trade-off Between Computational Cost And Performance according to the FMS ratio
FLOPs are calculated based on the test state(w/o the ATM). Each metric is rescaled according to the unit indicated.
}
\resizebox{0.9\columnwidth}{!}{
\begin{tabular}{@{}cccccc@{}}
\toprule
\multirow{3}{*}{FMS} & \multirow{3}{*}{\begin{tabular}[c]{@{}c@{}}FLOPs\\ (G)\end{tabular}} & \multicolumn{4}{c}{Arb. Scale Module}                             \\ \cmidrule(l){3-6} 
                     &                                                                      & \multicolumn{2}{c}{MSE ($\times$1e-5)}         & \multicolumn{2}{c}{MAE ($\times$1e-3)}         \\ \cmidrule(l){3-6} 
                     &                                                                      & Velocity       & Level          & Velocity       & Level          \\ \midrule
2:1                  & \textbf{547}                                                         & 1.255          & \textbf{0.144} & 1.403          & 0.961          \\
5:1                  & 649                                                                  & 1.457          & 0.276          & 3.017          & 1.014          \\
11:1                 & 721                                                                  & \textbf{0.683} & 0.376          & \textbf{1.220} & \textbf{0.897} \\
None                 & 819                                                                  & 3.322          & 0.202          & 3.302          & 1.049          \\ \bottomrule
\end{tabular}
}
\label{table::split_eval}
\end{table}

\subsubsection{Qualitative Results}
To evaluate the arbitrary-scale downscaling performance of our model on out-of-train distributions, we demonstrate results from a downscaling to a $\times$50 scale.
We compare and visualize the results of the proposed method with bicubic, the Baseline, Non-split, and our proposal model. 
Baseline means a model that only uses the feature extractor and the ASM which is the same used in the image super-resolution domain. 
The non-split is a model that adds the PE and the ATM to the Baseline, and finally, our proposal model is a model that adds FMS to the Non-split.

The qualitative results can be seen in Fig.~\ref{fig:qualitative_results}. 
The red bounding box shows whether the details are expressed well, and the green bounding box shows whether the overall texture is predicted well. 
In the results in the green bounding box, the Bicubic shows checkerboard artifacts and fails to produce high-resolution results, while the Ours maintains a similar texture as the GT. This means that our model can produce extremely high-resolution HR$^{arb}$ with a continuous representation more reliably than interpolation methods like the Bicubic.
In the red bounding box, the Bicubic again fails to produce high resolution by losing all the detail.
The Baseline and the Non-split show better results but fail to predict some detail. 
Finally, Ours shows the best results, reproducing details perfectly without over/under-estimating. This confirms that focusing on individual characteristics with the FMS and using the PE to capture better patterns that appear in a particular geographic region is an effective way to generate detail. 

\begin{figure*}[t!]
  \centering
  \includegraphics[width=0.9\linewidth]{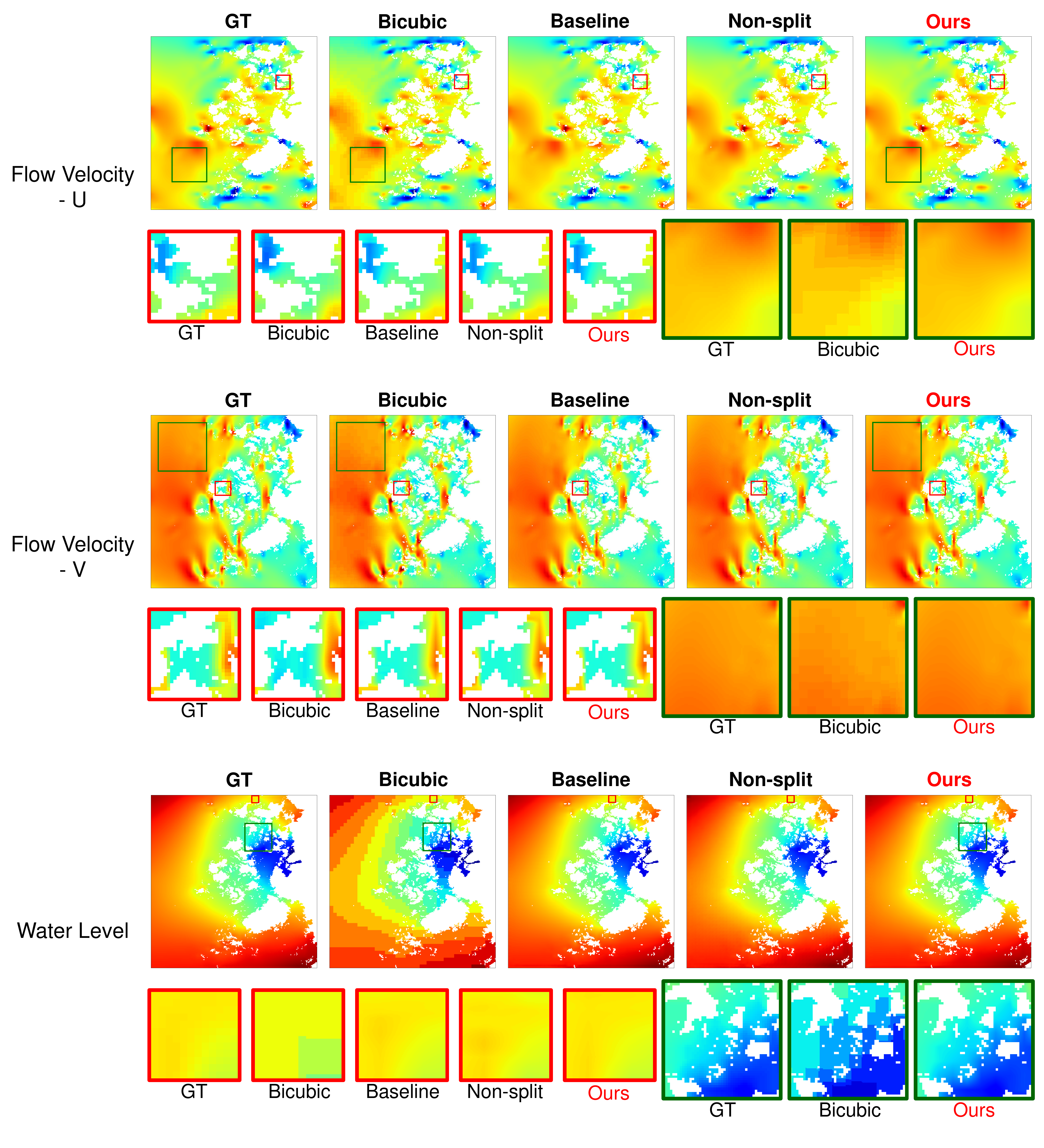}
  \vspace{-0.5cm}
  \caption{Visualization of the $\times$50 scale downscaling results. Green bounding boxes show overall texture and red bounding boxes show details of prediction. Since there is no huge difference between the INR-based methods(Baseline, Non-split, Ours) in texture, We only visualized Ours's result for the green bounding box between all results from the INR-based methods.}
  \label{fig:qualitative_results}
  \vspace{-0.3cm}
\end{figure*}

\section{Conclusion}
In this study, we proposed a novel framework for arbitrary-scale downscaling by taking the INR-based method used in other domains. 
For this framework, we proposed three components that take into account the characteristics of tidal current data produced by numerical modeling: an auxiliary train module, feature map splitting, and fusing a positional encoding.
Through the ablation study, the effectiveness of each component is evaluated. 
The proposed framework is not only capable of learning from scratch by applying an additional module but also reduces the computational cost by using the FMS.
Furthermore, we examined the results of downscaling to an out-of-train distribution at an $\times$50 scale, verifying that we can achieve stable, high-resolution large-scale prediction by learning continuous representations, even capturing fine details. 

\section{Data availity}
The authors do not have permission to share data. 

\section{Acknowledgements}
This work was supported by the 2022 Research Fund of the University of Seoul under project PID 202204281007.
 
The authors acknowledge the Korea Institute of Ocean Science \& Technology (KIOST) for generously sharing their data.

{
\bibliographystyle{model2-name}
\bibliography{egbib}

\begin{thebibliography}{32}
\expandafter\ifx\csname natexlab\endcsname\relax\def\natexlab#1{#1}\fi
\providecommand{\url}[1]{\texttt{#1}}
\providecommand{\href}[2]{#2}
\providecommand{\path}[1]{#1}
\providecommand{\DOIprefix}{doi:}
\providecommand{\ArXivprefix}{arXiv:}
\providecommand{\URLprefix}{URL: }
\providecommand{\Pubmedprefix}{pmid:}
\providecommand{\doi}[1]{\href{http://dx.doi.org/#1}{\path{#1}}}
\providecommand{\Pubmed}[1]{\href{pmid:#1}{\path{#1}}}
\providecommand{\bibinfo}[2]{#2}
\ifx\xfnm\relax \def\xfnm[#1]{\unskip,\space#1}\fi
%Type = Article
\bibitem[{Accarino et~al.(2021)Accarino, Chiarelli, Immorlano, Aloisi, Gatto
  and Aloisio}]{MSG-GAN-SD_2021}
\bibinfo{author}{Accarino, G.}, \bibinfo{author}{Chiarelli, M.},
  \bibinfo{author}{Immorlano, F.}, \bibinfo{author}{Aloisi, V.},
  \bibinfo{author}{Gatto, A.}, \bibinfo{author}{Aloisio, G.},
  \bibinfo{year}{2021}.
\newblock \bibinfo{title}{Msg-gan-sd: A multi-scale gradients gan for
  statistical downscaling of 2-meter temperature over the euro-cordex domain}.
\newblock \bibinfo{journal}{AI} \bibinfo{volume}{2}, \bibinfo{pages}{600--620}.
%Type = Inproceedings
\bibitem[{Agustsson and Timofte(2017)}]{Agustsson_2017_CVPRW_DIV2K}
\bibinfo{author}{Agustsson, E.}, \bibinfo{author}{Timofte, R.},
  \bibinfo{year}{2017}.
\newblock \bibinfo{title}{Ntire 2017 challenge on single image
  super-resolution: Dataset and study}, in: \bibinfo{booktitle}{The IEEE
  Conference on Computer Vision and Pattern Recognition (CVPR) Workshops}.
%Type = Article
\bibitem[{Baevski et~al.(2020)Baevski, Zhou, Mohamed and
  Auli}]{baevski2020wav2vec_cpe}
\bibinfo{author}{Baevski, A.}, \bibinfo{author}{Zhou, Y.},
  \bibinfo{author}{Mohamed, A.}, \bibinfo{author}{Auli, M.},
  \bibinfo{year}{2020}.
\newblock \bibinfo{title}{wav2vec 2.0: A framework for self-supervised learning
  of speech representations}.
\newblock \bibinfo{journal}{Advances in neural information processing systems}
  \bibinfo{volume}{33}, \bibinfo{pages}{12449--12460}.
%Type = Techreport
\bibitem[{Blain(1998)}]{blain1998coastal}
\bibinfo{author}{Blain, C.A.}, \bibinfo{year}{1998}.
\newblock \bibinfo{title}{Coastal tide prediction using the ADCIRC-2DDI
  hydrodynamic finite element model: Model validation and sensitivity analyses
  in the Southern North Sea/English Channel}.
\newblock \bibinfo{type}{Technical Report}. Naval Research Lab Stennis Space
  Center MS Coastal and Semi-Enclosed Seas~….
%Type = Inproceedings
\bibitem[{Cao et~al.(2023)Cao, Wang, Xian, Li, Ni, Pi, Zhang, Zhang, Timofte
  and Van~Gool}]{cao2023ciaosr}
\bibinfo{author}{Cao, J.}, \bibinfo{author}{Wang, Q.}, \bibinfo{author}{Xian,
  Y.}, \bibinfo{author}{Li, Y.}, \bibinfo{author}{Ni, B.}, \bibinfo{author}{Pi,
  Z.}, \bibinfo{author}{Zhang, K.}, \bibinfo{author}{Zhang, Y.},
  \bibinfo{author}{Timofte, R.}, \bibinfo{author}{Van~Gool, L.},
  \bibinfo{year}{2023}.
\newblock \bibinfo{title}{Ciaosr: Continuous implicit attention-in-attention
  network for arbitrary-scale image super-resolution}, in:
  \bibinfo{booktitle}{Proceedings of the IEEE/CVF Conference on Computer Vision
  and Pattern Recognition}, pp. \bibinfo{pages}{1796--1807}.
%Type = Inproceedings
\bibitem[{Chen et~al.(2021a)Chen, Wang, Guo, Xu, Deng, Liu, Ma, Xu, Xu and
  Gao}]{chen2021pre_CVPR}
\bibinfo{author}{Chen, H.}, \bibinfo{author}{Wang, Y.}, \bibinfo{author}{Guo,
  T.}, \bibinfo{author}{Xu, C.}, \bibinfo{author}{Deng, Y.},
  \bibinfo{author}{Liu, Z.}, \bibinfo{author}{Ma, S.}, \bibinfo{author}{Xu,
  C.}, \bibinfo{author}{Xu, C.}, \bibinfo{author}{Gao, W.},
  \bibinfo{year}{2021}a.
\newblock \bibinfo{title}{Pre-trained image processing transformer}, in:
  \bibinfo{booktitle}{Proceedings of the IEEE/CVF conference on computer vision
  and pattern recognition}, pp. \bibinfo{pages}{12299--12310}.
%Type = Inproceedings
\bibitem[{Chen et~al.(2023)Chen, Wang, Zhou, Qiao and Dong}]{chen2023hat_CVPR}
\bibinfo{author}{Chen, X.}, \bibinfo{author}{Wang, X.}, \bibinfo{author}{Zhou,
  J.}, \bibinfo{author}{Qiao, Y.}, \bibinfo{author}{Dong, C.},
  \bibinfo{year}{2023}.
\newblock \bibinfo{title}{Activating more pixels in image super-resolution
  transformer}, in: \bibinfo{booktitle}{Proceedings of the IEEE/CVF Conference
  on Computer Vision and Pattern Recognition (CVPR)}, pp.
  \bibinfo{pages}{22367--22377}.
%Type = Inproceedings
\bibitem[{Chen et~al.(2021b)Chen, Liu and Wang}]{LIIF_Chen_2021_CVPR}
\bibinfo{author}{Chen, Y.}, \bibinfo{author}{Liu, S.}, \bibinfo{author}{Wang,
  X.}, \bibinfo{year}{2021}b.
\newblock \bibinfo{title}{Learning continuous image representation with local
  implicit image function}, in: \bibinfo{booktitle}{Proceedings of the IEEE/CVF
  Conference on Computer Vision and Pattern Recognition (CVPR)}, pp.
  \bibinfo{pages}{8628--8638}.
%Type = Article
\bibitem[{Cheng et~al.(2020)Cheng, Kuang, Shen, Liu, Tan and Liu}]{LesLap_2020}
\bibinfo{author}{Cheng, J.}, \bibinfo{author}{Kuang, Q.},
  \bibinfo{author}{Shen, C.}, \bibinfo{author}{Liu, J.}, \bibinfo{author}{Tan,
  X.}, \bibinfo{author}{Liu, W.}, \bibinfo{year}{2020}.
\newblock \bibinfo{title}{Reslap: Generating high-resolution climate prediction
  through image super-resolution}.
\newblock \bibinfo{journal}{IEEE Access} \bibinfo{volume}{8},
  \bibinfo{pages}{39623--39634}.
%Type = Article
\bibitem[{Chu et~al.(2021)Chu, Tian, Zhang, Wang, Wei, Xia and
  Shen}]{chu2021conditional}
\bibinfo{author}{Chu, X.}, \bibinfo{author}{Tian, Z.}, \bibinfo{author}{Zhang,
  B.}, \bibinfo{author}{Wang, X.}, \bibinfo{author}{Wei, X.},
  \bibinfo{author}{Xia, H.}, \bibinfo{author}{Shen, C.}, \bibinfo{year}{2021}.
\newblock \bibinfo{title}{Conditional positional encodings for vision
  transformers}.
\newblock \bibinfo{journal}{arXiv preprint arXiv:2102.10882} .
%Type = Inproceedings
\bibitem[{Deng et~al.(2009)Deng, Dong, Socher, Li, Li and
  Fei-Fei}]{deng2009imagenet}
\bibinfo{author}{Deng, J.}, \bibinfo{author}{Dong, W.},
  \bibinfo{author}{Socher, R.}, \bibinfo{author}{Li, L.J.},
  \bibinfo{author}{Li, K.}, \bibinfo{author}{Fei-Fei, L.},
  \bibinfo{year}{2009}.
\newblock \bibinfo{title}{Imagenet: A large-scale hierarchical image database},
  in: \bibinfo{booktitle}{2009 IEEE conference on computer vision and pattern
  recognition}, \bibinfo{organization}{Ieee}. pp. \bibinfo{pages}{248--255}.
%Type = Inproceedings
\bibitem[{Dosovitskiy et~al.()Dosovitskiy, Beyer, Kolesnikov, Weissenborn,
  Zhai, Unterthiner, Dehghani, Minderer, Heigold, Gelly et~al.}]{ViT_2021}
\bibinfo{author}{Dosovitskiy, A.}, \bibinfo{author}{Beyer, L.},
  \bibinfo{author}{Kolesnikov, A.}, \bibinfo{author}{Weissenborn, D.},
  \bibinfo{author}{Zhai, X.}, \bibinfo{author}{Unterthiner, T.},
  \bibinfo{author}{Dehghani, M.}, \bibinfo{author}{Minderer, M.},
  \bibinfo{author}{Heigold, G.}, \bibinfo{author}{Gelly, S.}, et~al., .
\newblock \bibinfo{title}{An image is worth 16x16 words: Transformers for image
  recognition at scale}, in: \bibinfo{booktitle}{International Conference on
  Learning Representations}.
%Type = Article
\bibitem[{Geiss et~al.(2022)Geiss, Silva and Hardin}]{geiss2022downscaling}
\bibinfo{author}{Geiss, A.}, \bibinfo{author}{Silva, S.J.},
  \bibinfo{author}{Hardin, J.C.}, \bibinfo{year}{2022}.
\newblock \bibinfo{title}{Downscaling atmospheric chemistry simulations with
  physically consistent deep learning}.
\newblock \bibinfo{journal}{Geoscientific Model Development}
  \bibinfo{volume}{15}, \bibinfo{pages}{6677--6694}.
%Type = Misc
\bibitem[{Giorgi and Mearns(1999)}]{giorgi1999dynamicaldownscaling}
\bibinfo{author}{Giorgi, F.}, \bibinfo{author}{Mearns, L.O.},
  \bibinfo{year}{1999}.
\newblock \bibinfo{title}{Introduction to special section: Regional climate
  modeling revisited}.
%Type = Article
\bibitem[{Gomes et~al.(2015)Gomes, Neves, Kenov, Campuzano and
  Pinto}]{gomes2015tidemohid}
\bibinfo{author}{Gomes, N.}, \bibinfo{author}{Neves, R.},
  \bibinfo{author}{Kenov, I.A.}, \bibinfo{author}{Campuzano, F.J.},
  \bibinfo{author}{Pinto, L.}, \bibinfo{year}{2015}.
\newblock \bibinfo{title}{Tide and tidal currents in the cape verde
  archipelago}.
\newblock \bibinfo{journal}{Revista de Gest{\~a}o Costeira Integrada-Journal of
  Integrated Coastal Zone Management} \bibinfo{volume}{15},
  \bibinfo{pages}{395--408}.
%Type = Inproceedings
\bibitem[{Groenke et~al.(2020)Groenke, Madaus and
  Monteleoni}]{groenke2020climalign}
\bibinfo{author}{Groenke, B.}, \bibinfo{author}{Madaus, L.},
  \bibinfo{author}{Monteleoni, C.}, \bibinfo{year}{2020}.
\newblock \bibinfo{title}{Climalign: Unsupervised statistical downscaling of
  climate variables via normalizing flows}, in: \bibinfo{booktitle}{Proceedings
  of the 10th International Conference on Climate Informatics}, pp.
  \bibinfo{pages}{60--66}.
%Type = Article
\bibitem[{Harris et~al.(2022)Harris, McRae, Chantry, Dueben and
  Palmer}]{harris2022generative}
\bibinfo{author}{Harris, L.}, \bibinfo{author}{McRae, A.T.},
  \bibinfo{author}{Chantry, M.}, \bibinfo{author}{Dueben, P.D.},
  \bibinfo{author}{Palmer, T.N.}, \bibinfo{year}{2022}.
\newblock \bibinfo{title}{A generative deep learning approach to stochastic
  downscaling of precipitation forecasts}.
\newblock \bibinfo{journal}{Journal of Advances in Modeling Earth Systems}
  \bibinfo{volume}{14}, \bibinfo{pages}{e2022MS003120}.
%Type = Inproceedings
\bibitem[{Hu et~al.(2019)Hu, Mu, Zhang, Wang, Tan and Sun}]{hu2019metasr}
\bibinfo{author}{Hu, X.}, \bibinfo{author}{Mu, H.}, \bibinfo{author}{Zhang,
  X.}, \bibinfo{author}{Wang, Z.}, \bibinfo{author}{Tan, T.},
  \bibinfo{author}{Sun, J.}, \bibinfo{year}{2019}.
\newblock \bibinfo{title}{Meta-sr: A magnification-arbitrary network for
  super-resolution}, in: \bibinfo{booktitle}{Proceedings of the IEEE/CVF
  conference on computer vision and pattern recognition}, pp.
  \bibinfo{pages}{1575--1584}.
%Type = Inproceedings
\bibitem[{Lai et~al.(2017)Lai, Huang, Ahuja and Yang}]{LapSRN_2017}
\bibinfo{author}{Lai, W.S.}, \bibinfo{author}{Huang, J.B.},
  \bibinfo{author}{Ahuja, N.}, \bibinfo{author}{Yang, M.H.},
  \bibinfo{year}{2017}.
\newblock \bibinfo{title}{Deep laplacian pyramid networks for fast and accurate
  super-resolution}, in: \bibinfo{booktitle}{Proceedings of the IEEE conference
  on computer vision and pattern recognition}, pp. \bibinfo{pages}{624--632}.
%Type = Inproceedings
\bibitem[{Lee and Jin(2022)}]{lee2022lte}
\bibinfo{author}{Lee, J.}, \bibinfo{author}{Jin, K.H.}, \bibinfo{year}{2022}.
\newblock \bibinfo{title}{Local texture estimator for implicit representation
  function}, in: \bibinfo{booktitle}{Proceedings of the IEEE/CVF conference on
  computer vision and pattern recognition}, pp. \bibinfo{pages}{1929--1938}.
%Type = Inproceedings
\bibitem[{Lim et~al.(2017)Lim, Son, Kim, Nah and Mu~Lee}]{EDSR_2017}
\bibinfo{author}{Lim, B.}, \bibinfo{author}{Son, S.}, \bibinfo{author}{Kim,
  H.}, \bibinfo{author}{Nah, S.}, \bibinfo{author}{Mu~Lee, K.},
  \bibinfo{year}{2017}.
\newblock \bibinfo{title}{Enhanced deep residual networks for single image
  super-resolution}, in: \bibinfo{booktitle}{Proceedings of the IEEE conference
  on computer vision and pattern recognition workshops}, pp.
  \bibinfo{pages}{136--144}.
%Type = Inproceedings
\bibitem[{Mescheder et~al.(2019)Mescheder, Oechsle, Niemeyer, Nowozin and
  Geiger}]{mescheder2019occupancy}
\bibinfo{author}{Mescheder, L.}, \bibinfo{author}{Oechsle, M.},
  \bibinfo{author}{Niemeyer, M.}, \bibinfo{author}{Nowozin, S.},
  \bibinfo{author}{Geiger, A.}, \bibinfo{year}{2019}.
\newblock \bibinfo{title}{Occupancy networks: Learning 3d reconstruction in
  function space}, in: \bibinfo{booktitle}{Proceedings of the IEEE/CVF
  conference on computer vision and pattern recognition}, pp.
  \bibinfo{pages}{4460--4470}.
%Type = Article
\bibitem[{Mildenhall et~al.(2021)Mildenhall, Srinivasan, Tancik, Barron,
  Ramamoorthi and Ng}]{mildenhall2021nerf}
\bibinfo{author}{Mildenhall, B.}, \bibinfo{author}{Srinivasan, P.P.},
  \bibinfo{author}{Tancik, M.}, \bibinfo{author}{Barron, J.T.},
  \bibinfo{author}{Ramamoorthi, R.}, \bibinfo{author}{Ng, R.},
  \bibinfo{year}{2021}.
\newblock \bibinfo{title}{Nerf: Representing scenes as neural radiance fields
  for view synthesis}.
\newblock \bibinfo{journal}{Communications of the ACM} \bibinfo{volume}{65},
  \bibinfo{pages}{99--106}.
%Type = Article
\bibitem[{Nachtane et~al.(2020)Nachtane, Tarfaoui, Goda and
  Rouway}]{nachtane2020tidalcurrent_turbine}
\bibinfo{author}{Nachtane, M.}, \bibinfo{author}{Tarfaoui, M.},
  \bibinfo{author}{Goda, I.}, \bibinfo{author}{Rouway, M.},
  \bibinfo{year}{2020}.
\newblock \bibinfo{title}{A review on the technologies, design considerations
  and numerical models of tidal current turbines}.
\newblock \bibinfo{journal}{Renewable Energy} \bibinfo{volume}{157},
  \bibinfo{pages}{1274--1288}.
%Type = Article
\bibitem[{Nguyen et~al.(2023)Nguyen, Brandstetter, Kapoor, Gupta and
  Grover}]{ClimaX_2023}
\bibinfo{author}{Nguyen, T.}, \bibinfo{author}{Brandstetter, J.},
  \bibinfo{author}{Kapoor, A.}, \bibinfo{author}{Gupta, J.K.},
  \bibinfo{author}{Grover, A.}, \bibinfo{year}{2023}.
\newblock \bibinfo{title}{Climax: A foundation model for weather and climate}.
\newblock \bibinfo{journal}{arXiv preprint arXiv:2301.10343} .
%Type = Inproceedings
\bibitem[{Pak et~al.(2023)Pak, Lee and Jin}]{pak2023btc}
\bibinfo{author}{Pak, B.}, \bibinfo{author}{Lee, J.}, \bibinfo{author}{Jin,
  K.H.}, \bibinfo{year}{2023}.
\newblock \bibinfo{title}{B-spline texture coefficients estimator for screen
  content image super-resolution}, in: \bibinfo{booktitle}{Proceedings of the
  IEEE/CVF Conference on Computer Vision and Pattern Recognition}, pp.
  \bibinfo{pages}{10062--10071}.
%Type = Inproceedings
\bibitem[{Shi et~al.(2016)Shi, Caballero, Husz{\'a}r, Totz, Aitken, Bishop,
  Rueckert and Wang}]{espcn_2016}
\bibinfo{author}{Shi, W.}, \bibinfo{author}{Caballero, J.},
  \bibinfo{author}{Husz{\'a}r, F.}, \bibinfo{author}{Totz, J.},
  \bibinfo{author}{Aitken, A.P.}, \bibinfo{author}{Bishop, R.},
  \bibinfo{author}{Rueckert, D.}, \bibinfo{author}{Wang, Z.},
  \bibinfo{year}{2016}.
\newblock \bibinfo{title}{Real-time single image and video super-resolution
  using an efficient sub-pixel convolutional neural network}, in:
  \bibinfo{booktitle}{Proceedings of the IEEE conference on computer vision and
  pattern recognition}, pp. \bibinfo{pages}{1874--1883}.
%Type = Article
\bibitem[{Sitzmann et~al.(2020)Sitzmann, Martel, Bergman, Lindell and
  Wetzstein}]{sitzmann2020siren}
\bibinfo{author}{Sitzmann, V.}, \bibinfo{author}{Martel, J.},
  \bibinfo{author}{Bergman, A.}, \bibinfo{author}{Lindell, D.},
  \bibinfo{author}{Wetzstein, G.}, \bibinfo{year}{2020}.
\newblock \bibinfo{title}{Implicit neural representations with periodic
  activation functions}.
\newblock \bibinfo{journal}{Advances in neural information processing systems}
  \bibinfo{volume}{33}, \bibinfo{pages}{7462--7473}.
%Type = Article
\bibitem[{Thieler et~al.(2000)Thieler, Pilkey~Jr, Young, Bush and
  Chai}]{thieler2000coastalengineering}
\bibinfo{author}{Thieler, E.R.}, \bibinfo{author}{Pilkey~Jr, O.H.},
  \bibinfo{author}{Young, R.S.}, \bibinfo{author}{Bush, D.M.},
  \bibinfo{author}{Chai, F.}, \bibinfo{year}{2000}.
\newblock \bibinfo{title}{The use of mathematical models to predict beach
  behavior for us coastal engineering: a critical review}.
\newblock \bibinfo{journal}{Journal of Coastal Research} ,
  \bibinfo{pages}{48--70}.
%Type = Inproceedings
\bibitem[{Vandal et~al.(2017)Vandal, Kodra, Ganguly, Michaelis, Nemani and
  Ganguly}]{vandal2017deepsd}
\bibinfo{author}{Vandal, T.}, \bibinfo{author}{Kodra, E.},
  \bibinfo{author}{Ganguly, S.}, \bibinfo{author}{Michaelis, A.},
  \bibinfo{author}{Nemani, R.}, \bibinfo{author}{Ganguly, A.R.},
  \bibinfo{year}{2017}.
\newblock \bibinfo{title}{Deepsd: Generating high resolution climate change
  projections through single image super-resolution}, in:
  \bibinfo{booktitle}{Proceedings of the 23rd acm sigkdd international
  conference on knowledge discovery and data mining}, pp.
  \bibinfo{pages}{1663--1672}.
%Type = Inproceedings
\bibitem[{Vaswani et~al.(2017)Vaswani, Shazeer, Parmar, Uszkoreit, Jones,
  Gomez, Kaiser and Polosukhin}]{Transformer_2017}
\bibinfo{author}{Vaswani, A.}, \bibinfo{author}{Shazeer, N.},
  \bibinfo{author}{Parmar, N.}, \bibinfo{author}{Uszkoreit, J.},
  \bibinfo{author}{Jones, L.}, \bibinfo{author}{Gomez, A.N.},
  \bibinfo{author}{Kaiser, L.u.}, \bibinfo{author}{Polosukhin, I.},
  \bibinfo{year}{2017}.
\newblock \bibinfo{title}{Attention is all you need}, in:
  \bibinfo{editor}{Guyon, I.}, \bibinfo{editor}{Luxburg, U.V.},
  \bibinfo{editor}{Bengio, S.}, \bibinfo{editor}{Wallach, H.},
  \bibinfo{editor}{Fergus, R.}, \bibinfo{editor}{Vishwanathan, S.},
  \bibinfo{editor}{Garnett, R.} (Eds.), \bibinfo{booktitle}{Advances in Neural
  Information Processing Systems}, \bibinfo{publisher}{Curran Associates, Inc.}
%Type = Article
\bibitem[{Zhang et~al.(2022)Zhang, Huang, Liu, Wang and Jin}]{zhang2022swinfir}
\bibinfo{author}{Zhang, D.}, \bibinfo{author}{Huang, F.}, \bibinfo{author}{Liu,
  S.}, \bibinfo{author}{Wang, X.}, \bibinfo{author}{Jin, Z.},
  \bibinfo{year}{2022}.
\newblock \bibinfo{title}{Swinfir: Revisiting the swinir with fast fourier
  convolution and improved training for image super-resolution}.
\newblock \bibinfo{journal}{arXiv preprint arXiv:2208.11247} .

\end{thebibliography}
}

\end{document}